\documentclass{article}

    \PassOptionsToPackage{numbers, compress}{natbib}
 \usepackage[preprint]{neurips_2026}

\usepackage[utf8]{inputenc} 
\usepackage[T1]{fontenc}    
\usepackage{hyperref}       
\usepackage{url}            
\usepackage{booktabs}       
\usepackage{amsfonts}       
\usepackage{nicefrac}       
\usepackage{microtype}      
\usepackage{xcolor}         
\usepackage{subfig}
\usepackage{amsmath} 
\usepackage{amssymb}  
\usepackage{graphics} 
\usepackage{epsfig} 
\usepackage{algorithm}
\usepackage{algorithmic}
\usepackage{wrapfig}

\newcommand{\vect}[1]{{\boldsymbol{#1}}}
\newcommand{\E}{\mathbb{E} }

\title{Control-Aware Command Generation for Asynchronous Control and Action Inference with Vision-Based Policies towards Fast Execution}

\author{%
  Takayuki~Osa\\
  Center for Advanced Intelligence Project\\
  RIKEN\\
  \texttt{takayuki.osa@riken.jp} \\
  \And
  Kazuki~Ota \\
  Graduate School of Engineering\\
  The University of Tokyo\\
  Center for Advanced Intelligence Project\\
  RIKEN \\
  \texttt{ota@mi.t.u-tokyo.ac.jp} \\
}

\begin{document}

\maketitle

\begin{abstract}
Vision-based control policies, including vision–language–action models, have enabled robots to perform increasingly complex tasks. However, their execution speed is often limited by the high computational cost of action inference. In this work, we investigate what can be achieved at the control level without modifying the underlying policy learning process. We introduce an asynchronous inference and control framework that enables fast task execution while maintaining smooth and stable robot motions. By formulating the generation of continuous target positions from discontinuous model predictions as a tracking control problem, our method makes asynchronous inference practical for real robots. We further accelerate execution through adaptive action skipping based on the variance of predicted positions. Experiments with Action Chunking Transformer and OpenVLA-OFT on a bimanual robot demonstrate substantial speedups while preserving high task success rates. 
\end{abstract}

\section{Introduction}
Vision-based control policies, including large-scale vision-language-action (VLA) models, have recently been applied to a wide range of robotic tasks~\cite{OXE}. Compared with approaches that explicitly estimate object poses, they simplify system design and can be trained from human demonstrations through imitation learning~\cite{Argall09,Osa18}. By leveraging visual and language inputs, such policies have the potential to make robot control more accessible and expand the range of robotic applications.

\begin{wrapfigure}{r}{9cm}
    \centering
    \vspace{-0.4cm}
    \includegraphics[width=0.6\columnwidth]{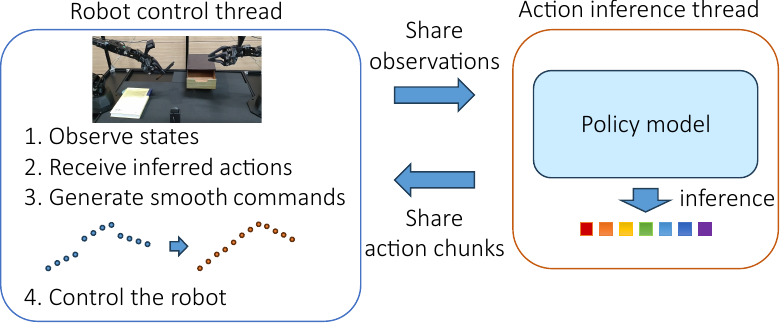}
    \caption{Illustration of the proposed framework with asynchronous robot control and action inference. A key challenge lies in generating commands to ensure physically plausible motions.}
    \label{fig:intro}
    \vspace{-0.4cm}
\end{wrapfigure}

In this work, we investigate how task execution can be accelerated without modifying the underlying policy learning process. As shown in Fig.~\ref{fig:intro}, we propose a framework that enables asynchronous action inference by formulating the generation of continuous target positions from discontinuous policy outputs as a tracking control problem. The proposed command generator produces smooth motions even when the policy outputs irregular action sequences. Furthermore, we accelerate execution through adaptive skipping of predicted actions based on the variance of predicted position sequences. We evaluate the method on a bimanual robot using both an 84M-parameter transformer policy and the 7B-parameter OpenVLA model, which are not diffusion or flow-matching-based. Experimental results demonstrate substantially faster task execution while preserving task performance.

\section{Related Work}
Vision-based control policies, including large-scale VLA models such as RT-X~\cite{OXE}, Octo~\cite{Octo23}, OpenVLA~\cite{Kim25}, $\pi_0$~\cite{Black25}, and $\pi_{0.5}$~\cite{pi25}, have recently achieved remarkable progress in robotic manipulation. Related approaches that predict actions from image observations and robot states have also been widely studied~\cite{Zhao23,Fu2024,Lee25,Zhao25b}, with ACT~\cite{Zhao23} being a representative example. A common limitation of these methods is the high computational cost of inference, which slows down task execution. 

Several recent studies have explored accelerating execution in imitation learning~\cite{Arachchige25,Guo25}. The method of~\cite{Arachchige25} exploits a high-fidelity robot controller to guide action generation, but its applicability to arbitrary robotic platforms remains unclear. Guo et al.~\cite{Guo25} accelerate execution by down-sampling demonstrations before training, yet do not explicitly address motion smoothness during execution. In contrast, our approach focuses on command generation at deployment time. Therefore, our contribution is orthogonal to that of Guo et al.~\cite{Guo25}, and the two methods can be readily combined.

A closely related effort is~\cite{Black25_chunk}, which tackles discontinuities caused by asynchronous inference through an inpainting formulation inspired by image generation~\cite{Lugmayr22} and diffusion guidance methods~\cite{Song23}. However, the approach is restricted to diffusion or flow-matching policies and is not directly applicable to architectures such as ACT~\cite{Zhao23} or OpenVLA~\cite{Kim25}. Moreover, because it relies on learned action generation, it does not provide explicit guarantees on constraints such as acceleration limits.


\section{Preliminary}
\subsection{Imitation learning}
The goal of imitation learning~\cite{Osa18} is to obtain a policy $\pi$ that reproduces the behavior of the expert policy $\pi^{\text{expert}}$.
This problem can be formulated as 
\begin{equation}
\pi^{*} = \arg \min_{\pi \in \Pi} D \left( \pi^{\text{expert}} || \pi\right) ,
\end{equation}
where $D \left( \pi^{\text{expert}} || \pi \right)$ is a function that quantifies the difference between the behaviors of $\pi$ and $\pi^{\text{expert}}$.
Given a dataset of the expert behaviors $\mathcal{D}^{\text{demo}}=\{ (\vect{s}_i, \vect{a}_i)\}^N_{i=1}$, the popular choice of $D \left( \pi^{\text{expert}} || \pi \right)$ is the expected KL divergence given by
\begin{align}
\E_{(\vect{s}, \vect{a}) \sim \mathcal{D}^{\text{demo}}} \left[ D_{\text{KL}} \left( \pi^{\text{expert}} || \pi \right) \right] = \int d^{\mathcal{D}^{\text{demo}}}(\vect{s}) \int \pi^{\text{expert}}(\vect{a}|\vect{s}) \log \frac{\pi^{\text{expert}}(\vect{a}|\vect{s})}{\pi(\vect{a}|\vect{s})} d\vect{a}d\vect{s},
\label{eq:kl}
\end{align}
where $d^{\mathcal{D}^{\text{demo}}}(\vect{s})$ is the empirical distribution of the expert demonstrations.
Minimizing the expected KL divergence in \eqref{eq:kl} is equivalent to maximizing the expected log likelihood $\E_{\vect{s}, \vect{a} \sim \mathcal{D}^{\text{demo}}} \left[ \log \pi(\vect{a}|\vect{s}) \right]$,
which is often referred to as behavioral cloning (BC).

\subsection{Action Chunking Transformer}
One of the widely used imitation learning frameworks for vision-based policies is ACT~\cite{Zhao23}.
In ACT, the policy is trained as a part of a conditional variational autoencoder (CVAE) ~\cite{Kingma14}.
The policy $\pi_{\vect{\theta}}$ with parameters $\vect{\theta}$  outputs a sequence of actions as
\begin{equation}
    \hat{\vect{a}}_{t:t+n_{\text{chunk}}-1} = \pi_{\vect{\theta}}(\vect{s}_t, \vect{z}_t)
\end{equation}
where $\vect{s}_t$ is the observation, $\vect{z}_t$ is the latent variable, and $n_{\text{chunk}}$ is the number of actions.
The latent variable is sampled from the posterior distribution modeled by an encoder during the training phase, and the latent variable is set as $\vect{z}=0$ when the policy is deployed. 

As the policy model generates a sequence of actions, an action at a given time step may be estimated multiple times.
In ACT, actions correspond to desired joint positions, $\vect{a}_t = [q^0_t, q^1_t, \ldots, q^d_t]$. The final action is obtained by aggregating multiple predictions through a weighted average~\cite{Zhao23}, 
\begin{equation} 
\vect{a}^{\text{act}}_t = \frac{\sum^{n-1}_{k=0} w_k \vect{a}^{t-k}_t} {\sum^{n-1}_{k=0} w_k}, 
\end{equation} 
where $w_k=\exp(c k)$, and $c<0$ controls the exponential weighting. While ACT and related methods achieve strong empirical performance~\cite{Zhao23,Fu2024,Lee25}, they provide no explicit guarantees on physical constraints such as velocity and acceleration limits.

\section{Proposed Framework}
\subsection{Control-Aware Command Generation}
\label{sec:cac}
In our framework, the actions inferred by the model are interpreted as the desired positions of the robot.
Given a sequence of desired positions $\{q^{\text{des}}_t, q^{\text{des}}_{t+1}, \ldots \}$ inferred by the model, we formulate the problem of generating a new sequence of target positions $\{q^{\text{targ}}_t, q^{\text{targ}}_{t+1}, \ldots \}$ that reliably tracks the inferred actions while satisfying high-speed constraints such as acceleration limits.
As commands to the robot, we use $\{q^{\text{targ}}_t, q^{\text{targ}}_{t+1}, \ldots \}$ rather than directly applying $\{q^{\text{des}}_t, q^{\text{des}}_{t+1}, \ldots \}$.
In practice, when asynchronous inference is employed, as illustrated in Fig.~\ref{fig:intro}, the sequence inferred by the model may be non-smooth, making it critical to regenerate a smooth sequence of target positions.

\begin{wrapfigure}{r}{6cm}
    \centering
    \includegraphics[width=0.4\columnwidth]{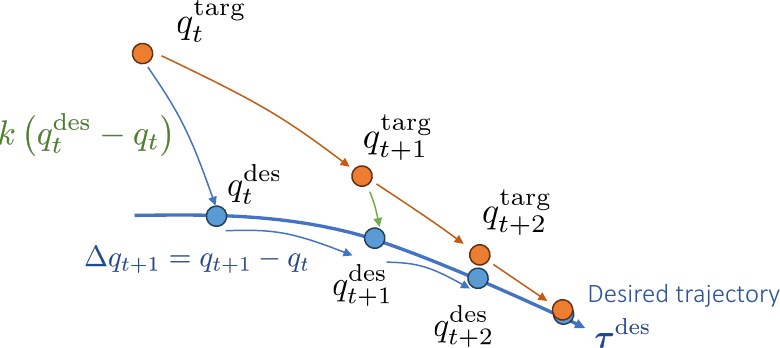}
    \caption{Illustration of control-aware command generation. 
    The target positions converge to the desired position while satisfying constraints such as acceleration limits.
    }
    \label{fig:cac}
    \vspace{-0.3cm}
\end{wrapfigure}

To achieve a stable tracking, we propose to compute the target position at time $t+1$
\begin{equation}
      q^{\text{targ}}_{t+1} = q^{\text{targ}}_{t} + (q^{\text{des}}_{t+1} - q^{\text{des}}_t) + k \cdot \text{sat}(q^{\text{des}}_t - q^{\text{targ}}_t, \delta),
\end{equation}
where the term $q^{\text{des}}_{t+1} - q^{\text{des}}_t$ can be interpreted as a feedforward control component, while $k \cdot \text{sat}(q^{\text{des}}_t - q^{\text{targ}}_t, \delta)$ serves as a feedback component, as illustrated in Fig.~\ref{fig:cac}.
The saturation function $\text{sat}(x,\delta)$ is defined as $\text{sat}(x,\delta)=\min(\max(x,-\delta),\delta)$.
This formulation for updating $q^{\text{targ}}_{t+1}$ is analogous to sliding mode control~\cite{Slotine91}, enabling stable tracking of a moving target.
The convergence of this type of control is discussed in Appendix~\ref{app:convergence}

In addition, we impose a constraint based on the maximum allowable acceleration.
When the upper and lower bounds of discrete acceleration are denoted by $a_{\text{max}}$ and $a_{\text{min}}$, respectively, $q^{\text{targ}}_{t+1}$ is computed as
\begin{equation}
      q^{\text{targ}}_{t+1} = q^{\text{targ}}_{t} + \text{clip}(\Delta \tilde{q}^{\text{targ}}_{t+1}, \Delta q^{\text{targ}}_{t} + a_{\text{min}}, \Delta q^{\text{targ}}_t + a_{\text{max}}),
      \label{eq:cac}
\end{equation}
where $a_{\text{min}} < 0 < a_{\text{max}}$ and
\begin{equation}
\Delta \tilde{q}^{\text{targ}}_{t+1} = \Delta {q}^{\text{des}}_{t+1} + k \cdot \text{sat}(q^{\text{des}}_t - q^{\text{targ}}_t, \delta).
\end{equation}
Herein, $\Delta {q}^{\text{targ}}_t = q^{\text{targ}}_{t} - q^{\text{targ}}_{t-1}$ and $\text{clip}(x, x_{\text{min}}, x_{\text{max}})$ is a clipping function.
We refer to the method that generates commands according to \eqref{eq:cac} as Control-Aware Command Generation (CAC).


\begin{figure}[tb]
      \centering
       \subfloat[A toy example of a sequence of desired positions with large gaps.]{%
    \includegraphics[width=0.4\columnwidth]{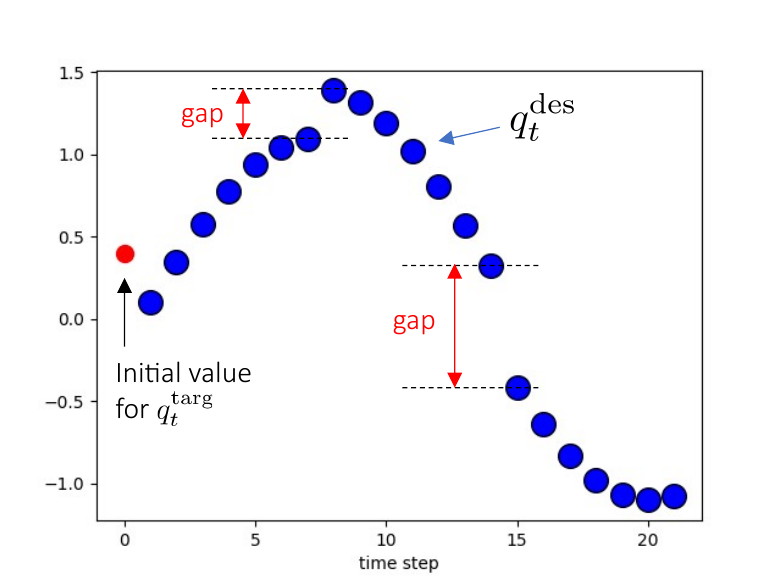}
    \label{fig:toy_setting}
  }
   \hspace{1cm}
  \subfloat[Commands generated with the proposed method.]{%
    \includegraphics[width=0.4\columnwidth]{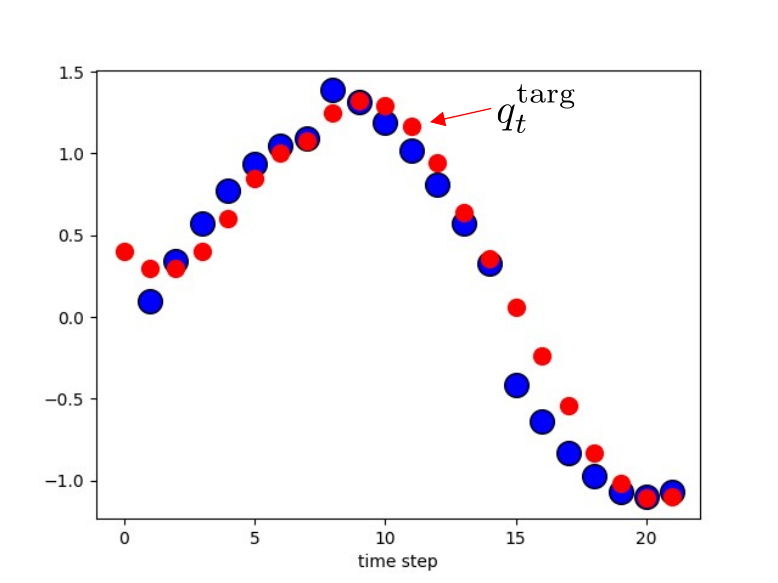}
    \vspace{-0.5cm}
    \label{fig:toy_command}
  }
  \caption{The blue points represent a sequence of desired positions, which correspond to $q^{\text{des}}_t$, and and the red points in (b) represent the commands generated with the proposed method,  $q^{\text{targ}}_t$. 
  }
  \label{fig:toy}
\end{figure}

An example of the behavior of the proposed method is shown in Fig.~\ref{fig:toy}.
In this toy example, the initial value of $q^{\text{targ}}$ is set to the initial position of the actual robot, while the initial value of $q^{\text{des}}$ is offset from that of $q^{\text{targ}}$.
Furthermore, the sequence of $q^{\text{des}}$ includes two gaps, as illustrated in Fig.~\ref{fig:toy_setting}.
As shown in Fig.~\ref{fig:toy_command}, the proposed method generates a sequence of points that follow the desired positions smoothly while satisfying the constraints.

\begin{figure}[t] 
\centering 
\begin{minipage}[t]{0.46\linewidth}
\vspace{0pt}
    \centering
    \includegraphics[width=\columnwidth]{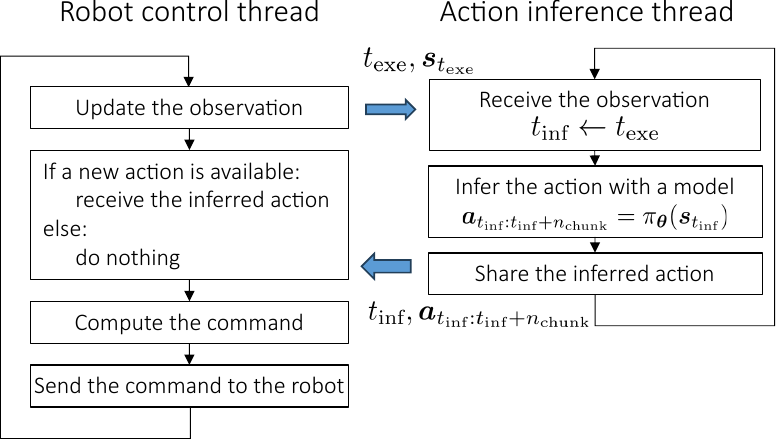}
    \caption{Threads for robot control and action inference.}
    \label{fig:multithreading}
\end{minipage} 
\hfill 
\begin{minipage}[t]{0.46\linewidth}
\vspace{0pt}
\centering
    \includegraphics[width=\columnwidth]{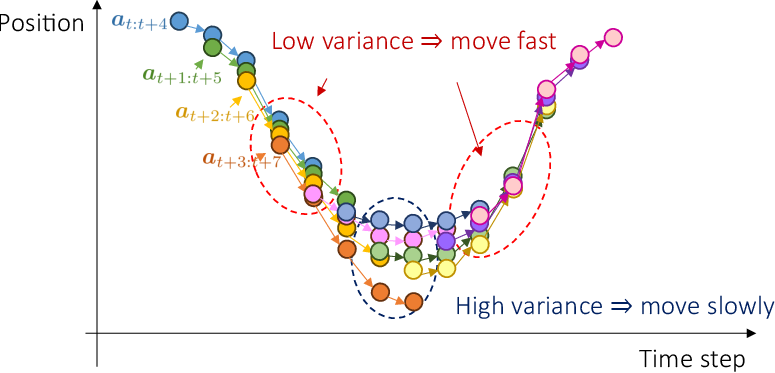}
    \caption{Illustration of the adaptive task execution based on the variance of the predicted actions. The colors indicate the time step when the action sequence was predicted.}
    \label{fig:traj_variance}
    \vspace{-0.3cm}
\end{minipage}
\vspace{-0.3cm}
\end{figure}

\subsection{Implementation with Multithreading}
\label{sec:multithread}

To achieve high control frequency and fast task execution, we separate the control loop and the inference loop into different threads, as shown in Fig.~\ref{fig:multithreading}, which enables asynchronous inference.
The control thread updates observations, shares them with the inference loop, and periodically checks for new action chunks.
If a new chunk is available, it is used. Otherwise, previously received data are used.
Since one inference step produces a sequence of $n_{\text{chunk}}$ actions, delays can be tolerated for up to $n_{\text{chunk}}$ steps.
Each observation and action chunk is tagged with a time stamp, denoted as $t_{\text{inf}}$ for inference, as illustrated in Fig.~\ref{fig:multithreading}.
Because the control loop runs faster, $t_{\text{exe}}$ is typically greater than $t_{\text{inf}}$ when actions are retrieved.

\subsection{Fast and Adaptive Task Execution}
\label{sec:adaptive}

While asynchronous action inference and control provide one means of achieving fast task execution, another straightforward approach is to skip certain steps within action chunks rather than executing all of them.
However, skipping introduces two key challenges.
First, it produces non-smooth action sequences. 
This can be mitigated using the control-aware command generation described in Section~\ref{sec:cac}, which smooths non-smooth desired positions $q^{\text{des}}_t$. 
Second, skipping reduces the reliability of older inferred actions, since it increases the gap between the inference step $t_{\text{inf}}$ and the execution step $t_{\text{exe}}$. 
To address this, we adaptively adjust the execution speed: the robot moves faster when the inferred actions are reliable and more slowly otherwise.

To evaluate the reliability of generated actions, we compute the variance of predictions across time steps. 
Specifically, given a set of actions for step $t$, denoted as $\mathcal{A}=\{ \vect{a}^t_t, \vect{a}^{t-1}_t, \ldots, \vect{a}^{t-n+1}_t \}$, 
the variance is defined as
\begin{equation}
    \text{Var}(\vect{a}_t) = \sum^D_{i=1} \left( \frac{1}{n} \sum^{n-1}_{k=0} (a^{t-k}_{t,i} - \mu_i)^2 \right),
\end{equation}
where $a^{t-k}_{t,i}$ is the $i$th element of $\vect{a}^{t-k}_t$, and $\mu_i = n^{-1} \sum^n_{k=1} a^{t-k}_{t,i}$ is its mean. 
In our framework, if $\text{Var}(\vect{a}_t)$ is smaller than a threshold $c_{\text{var}}$, the current command is skipped and the time step is updated as $t \leftarrow t+2$ instead of $t \leftarrow t+1$.

The intuition is illustrated in Fig.~\ref{fig:traj_variance}. 
When $\text{Var}(\vect{a}_t)$ is low, the difference between older and newer predictions is small, indicating reliable actions, so the robot can move faster. 
When $\text{Var}(\vect{a}_t)$ is high, the difference is large and the newer prediction significantly updates the action, so the robot should move more slowly to follow the updated inference. 


\section{Experiments}

\begin{figure}[tb]
\centering
\subfloat[The box handover task.]{%
    \includegraphics[width=0.24\columnwidth]{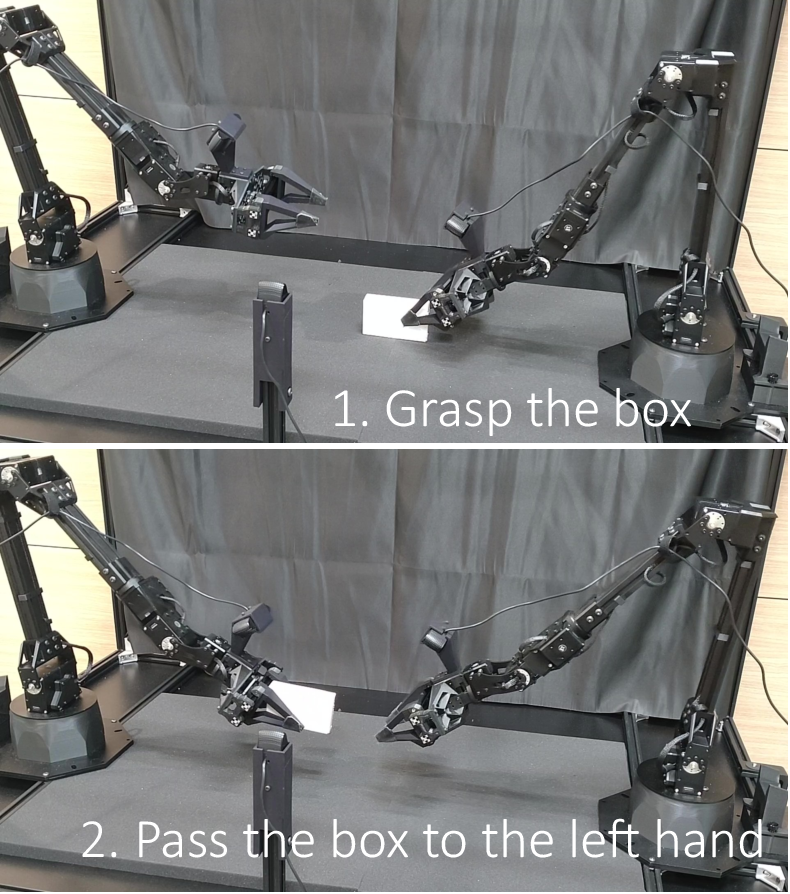}
    \label{fig:hand_over}
  }
   \hfill
  \subfloat[The case stacking task.]{%
    \includegraphics[width=0.24\columnwidth]{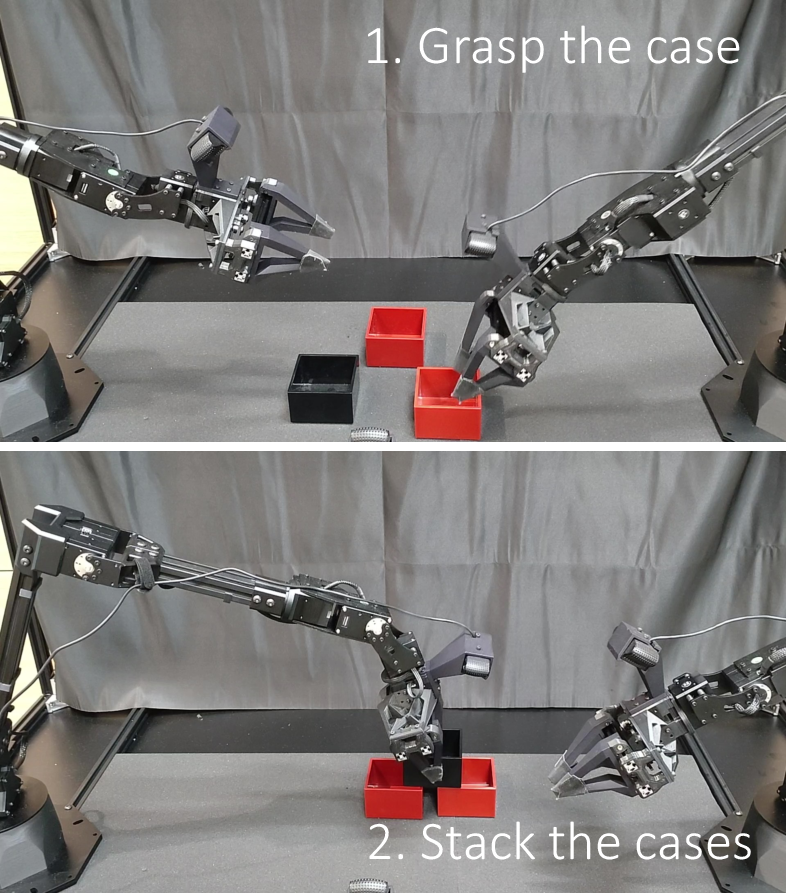}
    \label{fig:block_stack}
  }
  \hfill
\subfloat[The pen storage task.]{%
    \includegraphics[width=0.48\columnwidth]{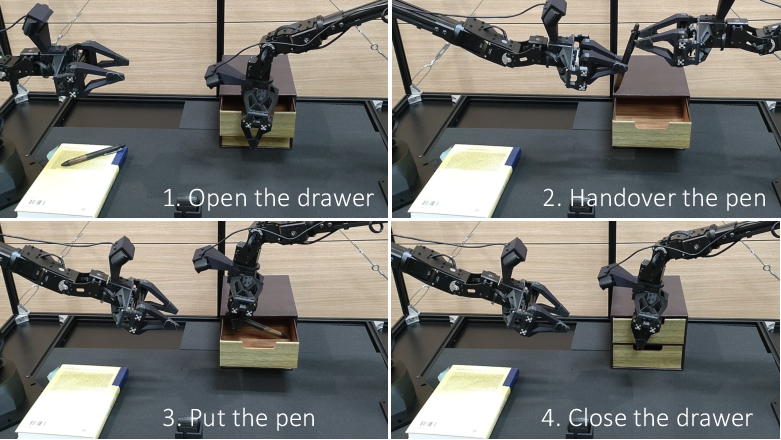}
    \label{fig:pen_storage}
  }
  \vspace{-0.1cm}
  \caption{Snapshots of the tasks for the real robot experiment.}
  \label{fig:exp_tasks}
\end{figure}

We evaluate the proposed control architecture on a stationary ALOHA system following~\cite{Zhao23}.
We conducted real-robot experiments with three tasks, shown in Figure~\ref{fig:exp_tasks}.
Models trained with three different random seeds were evaluated.
Each test trial was repeated 10 times per seed in the experiments.


To evaluate the speed and smoothness of the motion, we compute the discrete velocity $\Delta \vect{q}_{t}$, acceleration $\Delta^{2}\vect{q}_{t}$, and jerk $\Delta^{3}\vect{q}_{t}$ as
\begin{eqnarray}
\Delta\vect{q}_{t} &=& \vect{q}_{t} - \vect{q}_{t-1},  
\Delta^{2}\vect{q}_{t} = \Delta\vect{q}_{t} - \Delta\vect{q}_{t-1}, \nonumber\\
\Delta^{3}\vect{q}_{t} &=& \Delta^{2}\vect{q}_{t} - \Delta^{2}\vect{q}_{t-1}, \label{eq:jerk_approx}
\end{eqnarray}
where velocity, acceleration, and jerk correspond to the first-, second-, and third-order finite differences, respectively.

To examine the effect of the acceleration constraint, we visualize the maximum acceleration across all dimensions at each time step, defined as $\max_{i} \Delta^{2} q_{t,i}$, where $\Delta^{2} q_{t,i}$ denotes the $i$th element of $\Delta^{2}\vect{q}_{t}$.
Similarly, we analyze the maximum jerk across all dimensions at each time step, defined as $\max_{i} \Delta^{3} q_{t,i}$, where $\Delta^{3} q_{t,i}$ denotes the $i$th element of $\Delta^{3}\vect{q}_{t}$.

The main experiments were conducted using ACT~\cite{Zhao23}. 
To examine the effect across different policy models, part of the experiments were also performed with a fine-tuned version of OpenVLA~\cite{Kim25}, a VLA model. 
For fine-tuning OpenVLA, we followed the OpenVLA-OFT procedure~\cite{Kim25-openvlaoft}. 
It is worth noting that neither ACT nor OpenVLA-OFT is diffusion- or flow-matching-based, making the real-time chunking method of \citet{Black25_chunk} inapplicable.
A video of the experiments is available \href{https://www.youtube.com/watch?v=IX5TGaNot9A}{here}.

\subsection{Results of the Real Robot Experiment based on ACT}

\begin{table}[tb]
\centering
\caption{Success rate of the tasks with the real robot.}
\label{tbl:success_act}
\begin{tabular}{cccccc}
\toprule
Task &
\parbox{0.9cm}{\centering ACT +Sync} & 
\parbox{0.9cm}{\centering ACT +Async}& 
\parbox{1.2cm}{\centering ACT \\ +CAC +Async} & 
\parbox{1.2cm}{\centering ACT \\ +CAC +Async +Skip} & 
\parbox{1.5cm}{\centering ACT \\ +CAC +Async +Adaptive}\\
\midrule
\parbox{2.cm}{\centering Box handover} & \textbf{100.0\%} & \textbf{100.0\%} & \textbf{100.0\%} & 86.7 \% & \textbf{96.7\%}\\
\parbox{2.cm}{\centering Case stacking} & \textbf{100.0\%} & \textbf{100.0\%} & \textbf{100.0\%} & \textbf{100.0\%} & \textbf{100.0\%}\\
\parbox{2.cm}{\centering Pen storage} & 30.0\% & 63.3\% & 70.0\% & 0 \% & \textbf{83.3\%}\\
\bottomrule
\end{tabular}
\vspace{-0.4cm}
\end{table}

We evaluated the effects of CAC, asynchronous inference, and the proposed adaptive execution-speed control. ACT with synchronous and asynchronous execution are denoted by ACT+Sync and ACT+Async, respectively. Variants using CAC, adaptive speed control, and uniform command skipping are denoted by +CAC, +Adaptive, and +Skip.

\begin{figure}[tb]
\centering
  \subfloat[The norm of the discrete velocity $\left\|\Delta\vect{q}_{t}\right\|_2$.]{%
    \includegraphics[width=0.31\columnwidth]{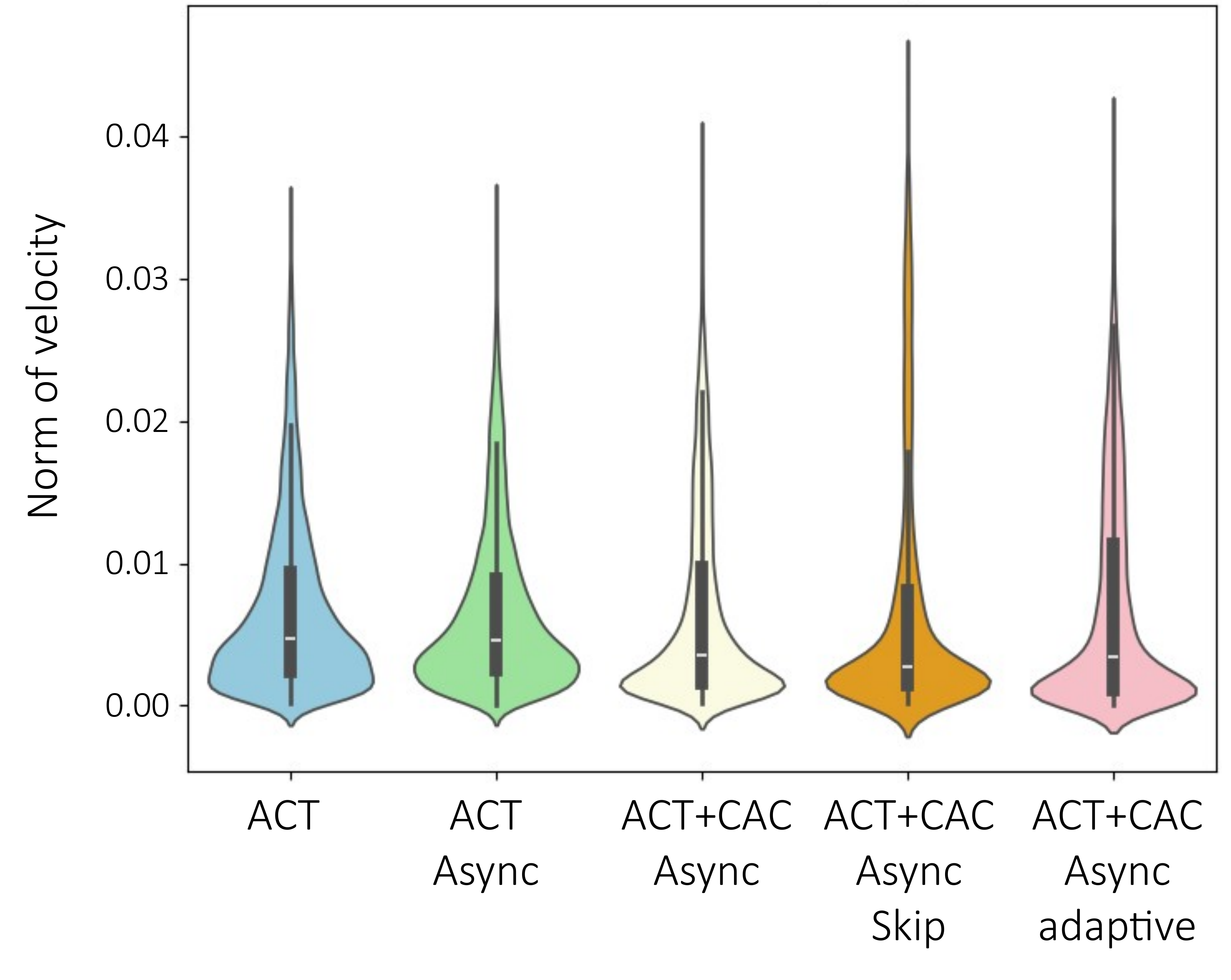}
    \label{fig:exp_vel_norm}
  }
    \hfill
       \subfloat[The discrete acceleration $\max_{i}\Delta^2 q_{t,i}$.]{%
    \includegraphics[width=0.31\columnwidth]{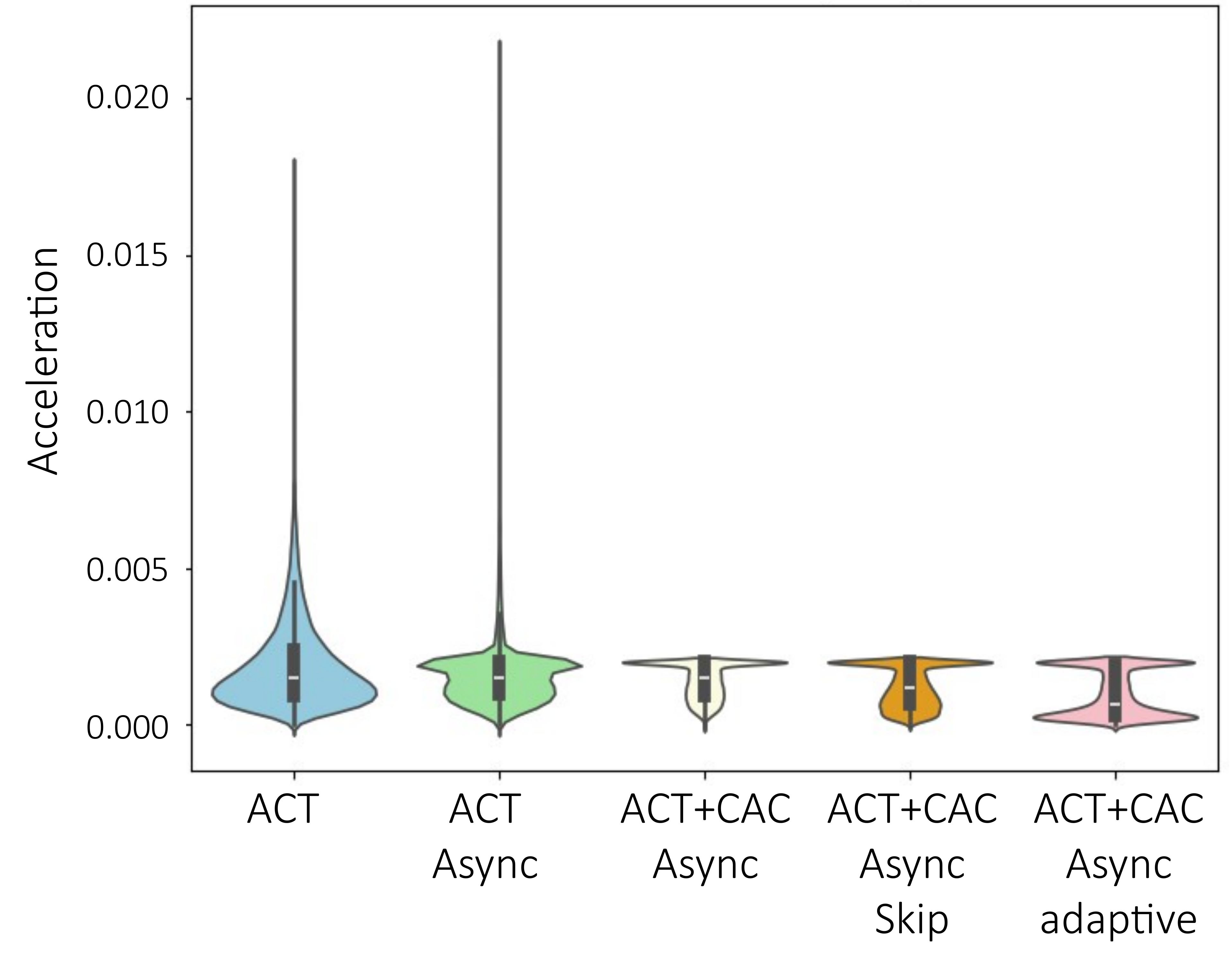}
    \label{fig:exp_acc_max}
  }
  \hfill
  \subfloat[The discrete jerk $\max_{i}\Delta^3 q_{t,i}$.]{%
    \includegraphics[width=0.31\columnwidth]{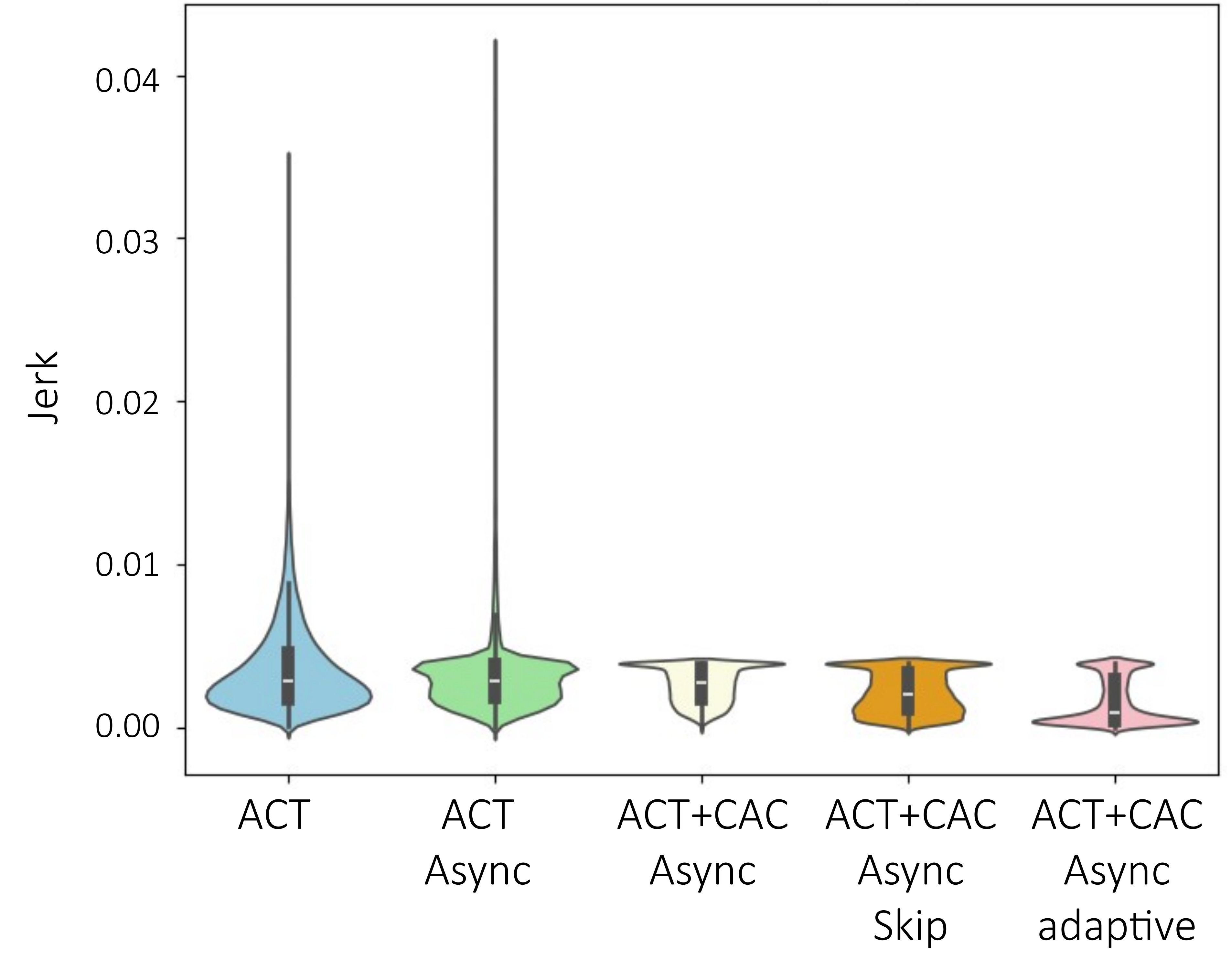}
    \label{fig:exp_jerk_max}
  }
  \caption{The velocity, acceleration and jerk of commands across control schemes.}
  \label{fig:exp_traj}
\vspace{-0.2cm}
\end{figure}



Figure~\ref{fig:exp_vel_norm} shows that ACT +CAC +Async +Adaptive achieved the highest execution speed. Although the discrete velocity distributions were similar between ACT+Sync and ACT+Async, asynchronous execution resulted in a faster control cycle and therefore higher actual motion speed. As shown in Figs.~\ref{fig:exp_acc_max} and~\ref{fig:exp_jerk_max}, CAC effectively limited acceleration and significantly reduced jerk, leading to smoother motions and improved vibration suppression. In contrast, asynchronous ACT without CAC produced larger accelerations and jerks, likely because actions were generated from delayed observations and executed at a faster control frequency. Table~\ref{tbl:success_act} shows that all methods achieved similar success rates on relatively simple tasks. 

\begin{table}[tb]
\caption{The task execution time. }
\label{tbl:time_act}
\begin{center}
\begin{tabular}{cccccc}
\toprule
Task &
\parbox{0.9cm}{\centering ACT +Sync} & 
\parbox{0.9cm}{\centering ACT +Async}& 
\parbox{1.2cm}{\centering ACT \\ +CAC +Async} & 
\parbox{1.2cm}{\centering ACT \\ +CAC +Async +Skip} & 
\parbox{1.5cm}{\centering ACT \\ +CAC +Async +Adaptive}\\
\midrule
\parbox{2.cm}{\centering Box handover} & 16.2$\pm$0.3  & 11.0$\pm$0.2 &11.1$\pm$0.1  & \textbf{7.2$\pm$0.3}  & 8.6$\pm$0.1 \\
\parbox{1.9cm}{\centering Case stacking} & 34.4$\pm$1.4  & 21.8$\pm$0.2  & 22.2$\pm$0.6 & \textbf{12.9$\pm$0.3} & 19.2$\pm$0.6   \\
\parbox{1.9cm}{\centering Pen storage} & 52.6$\pm$0.9 & \textbf{33.4 $\pm$1.2} & 38.6$\pm$1.0 & - & \textbf{34.4$\pm$0.8}\\
\bottomrule
\multicolumn{6}{l}{\parbox{8.0cm}{\footnotesize All values are in seconds. The mean and standard deviation are shown.}} 
\end{tabular}
\end{center}
\end{table}

However, Table~\ref{tbl:time_act} demonstrates that ACT +CAC +Async +Adaptive reduced task execution time to 53--63\% of ACT+Sync. The largest improvement was observed in the pen storage task, where the success rate increased from 30\% with ACT+Sync to 83\% with ACT+CAC+Async+Adaptive. In contrast, ACT+CAC+Async+Skip failed completely, indicating that uniform skipping degraded motion accuracy. Similar trends were observed in the box handover task. These results show that adaptive speed control can effectively balance execution speed and task success.

In the pen-storage task, asynchronous execution substantially improved performance even without CAC. We observed that slow motions during pen grasping often resulted in an unstable grip, causing failures in the subsequent handover to the right hand. Asynchronous execution significantly increased the speed of arm and gripper motions, resulting in a more stable grasp and more reliable handover. These observations provide a plausible explanation for the large performance gain from ACT+Sync to ACT+Async.

\subsection{Results of the Experiment based on OpenVLA-OFT}
To verify that CAC is applicable beyond ACT, we evaluated it with an OpenVLA model fine-tuned using OpenVLA-OFT on the pen storage task. We compared three settings: standard OpenVLA-OFT, OpenVLA-OFT+CAC, and OpenVLA-OFT+CAC with asynchronous inference. For asynchronous execution, we used the multi-threaded structure described in Section~\ref{sec:multithread} together with temporal ensembling~\cite{Zhao23}.

\begin{figure}[tb]
\centering
  \subfloat[The discrete acceleration of the commands $\max_{i}\Delta^2 q_{t,i}$.]{%
    \includegraphics[width=0.4\columnwidth]{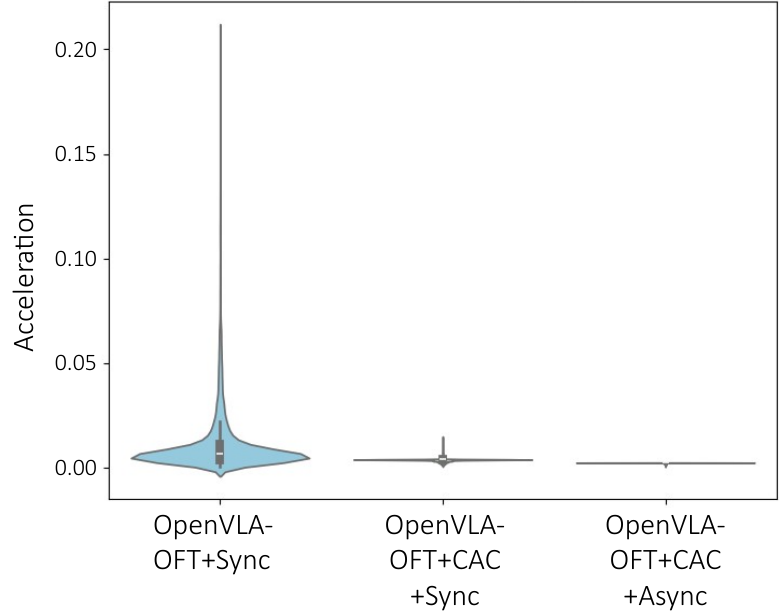}
    \label{fig:sim_acc}
  }
  \hfill
  \subfloat[The discrete jerk of the commands $\max_{i}\Delta^3 q_{t,i}$. ]{%
    \includegraphics[width=0.4\columnwidth]{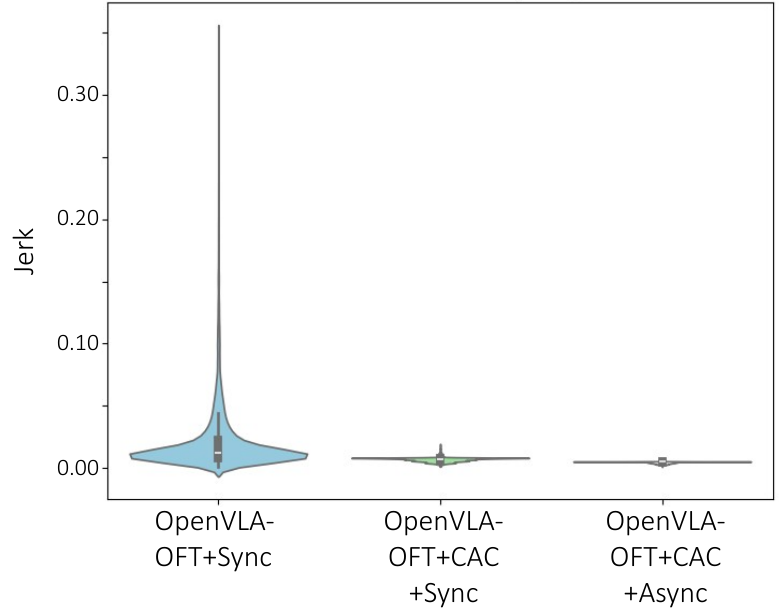}
    \label{fig:sim_jerk}
  }
  \caption{The acceleration and jerk in the pen storage task with the OpenVLA-OFT model.}
  \label{fig:oft_result}
  \end{figure}

Without CAC, asynchronous OpenVLA-OFT produced severe oscillations and could not reliably complete the task. Following the default OpenVLA-OFT setting, the chunk size was fixed at 25. Table~\ref{tbl:oft_result} shows that all evaluated methods achieved comparable success rates of roughly 80\%. However, CAC substantially improved trajectory smoothness. As illustrated in Fig.~\ref{fig:oft_result}, OpenVLA-OFT generated much larger acceleration and jerk values than ACT. While direct comparison is difficult because ACT employs temporal ensembling, applying OpenVLA-OFT outputs directly resulted in noticeable oscillations, particularly under asynchronous inference. 

\begin{table}[tb]
\centering
\caption{Results with OpenVLA-OFT-based model}
\label{tbl:oft_result}
\begin{center}
\begin{tabular}{cccc}
\toprule
Task &
\parbox{2.3cm}{\centering OpenVLA-OFT +Sync} & \parbox{2.3cm}{\centering OpenVLA-OFT +CAC +Sync} & \parbox{2.3cm}{\centering OpenVLA-OFT +CAC +Async} \\
\midrule
\parbox{3.cm}{\centering success rate} & \textbf{8/10} & \textbf{9/10} & \textbf{8/10} \\
\parbox{3.cm}{\centering episode duration [s]} & 68.2$\pm$0.1 & 68.2$\pm$0.1 & \textbf{59.0$\pm$0.1} \\
\parbox{3.2cm}{\centering task execution time [s]} & 49.0$\pm$0.2 & 52.6$\pm$1.6 & \textbf{44.7$\pm$2.5} \\
\bottomrule
\end{tabular}
\end{center}
\vspace{-0.4cm}
\end{table}

CAC effectively suppressed acceleration and jerk, enabling stable control in both synchronous and asynchronous settings. Asynchronous inference also reduced waiting time and shortened episode duration. The average inference waiting time decreased from approximately 18.5 seconds per episode in the synchronous setting to less than 0.8 seconds with asynchronous execution. Although OpenVLA-OFT+CAC+Async remained slower than ACT+CAC+Async due to the higher computational cost of OpenVLA-OFT inference, the results demonstrate that the proposed framework can effectively accelerate execution even for large VLA models.

\section{Conclusions}
We presented a framework for fast task execution with vision-based policies by enabling asynchronous action inference. 
We addressed the problem of generating action sequences that stably follow those produced by a policy model while satisfying high-speed constraints such as acceleration. 
Experimental results showed that our framework executes tasks faster than standard approaches. 
Through experiments with ACT and OpenVLA-OFT, we demonstrated that the proposed method reduces acceleration and jerk, thereby enabling the use of asynchronous inference in vision-based policies.
In the present experiments, we used the default parameters of OpenVLA-OFT without tuning, including the chunk size. 
As future work, we plan to conduct experiments with larger models and explore effective methods for them. 

In addition, we acknowledge that the current study does not include comparisons with generic smoothing methods or acceleration approaches designed for diffusion-based policies such as Real Time Chunking~\citep{Black25_chunk}.
A systematic comparison of execution-time acceleration strategies remains an important direction for future work.

More broadly, the significance of this work is not limited to the specific methodology for enabling asynchronous execution. 
Our results suggest that substantial improvements in execution speed and task performance can be achieved by appropriately designing the system that converts policy outputs into robot commands, without modifying the underlying policy model itself. 
This finding highlights the importance of considering the entire deployment pipeline of robot learning systems, including the execution layer, rather than focusing solely on policy learning algorithms.

\begin{ack}
Takayuki Osa was supported by JSPS KAKENHI Grant Number JP25K03176. Kazuki Ota was supported by JST SPRING, Grant Number JPMJSP2108.
\end{ack}


\bibliographystyle{unsrtnat}
\bibliography{neurips26wsbib}


\appendix



\section{Details of the Proposed Framework}

\subsection{Tracking Control in Continuous Time}
\label{app:convergence}
We consider the problem in continuous time. We denote the error between the desired joint position $q^{\text{des}}(t)$ and the target joint position $q^{\text{targ}}(t)$ as
\begin{equation}
    e(t) = q^{\text{des}}(t) - q^{\text{targ}}(t) .
\end{equation}
Herein, we consider the dynamics of the error function $\frac{1}{2}\left\|e\right\|^2_2$.
If the derivative of the error function satisfies
\begin{equation}
    \frac{d}{dt} \left(\frac{1}{2} \left\|e\right\|^2_2 \right) =  e \cdot \dot{e} =  - \frac{k}{2} e^2,
\end{equation}
the error function satisfies
\begin{equation}
\frac{1}{2} \left\| e(t) \right\|^2_2 = C \exp(-kt),
\end{equation}
where $C$ is a constant, and we obtain exponential convergence of the error.  
Similarly, if
\begin{equation}
    \dot{e}=  - K\cdot \text{sign}(e),
\end{equation}
where $K$ is a constant, the error decreases linearly, and finite-time convergence is achieved.

Based on the above observation, we aim to determine the target position at time $t$ that satisfies
\begin{equation}
    \dot{q}^{\text{targ}}(t) = \dot{q}^{\text{des}}(t) + k \cdot\text{sat}(q^{\text{des}}(t) - q^{\text{targ}}(t), \delta),
    \label{eq:com_cont_time}
\end{equation}
where $\text{sat}(x, \delta)$ is a saturation function defined as
\begin{align}
\text{sat}(x, \delta) = 
\left\{
\begin{array}{ll}
x & \text{if  } |x| < \delta,  \\
\delta \cdot\text{sign}(x) & \text{otherwise}, 
\end{array}
\right.
\end{align}
and $\delta$ is a constant.
When the derivative of the target position satisfies \eqref{eq:com_cont_time}, the error function decays linearly for $|e|>\delta$ and exponentially otherwise.

Based on the discussion above, we designed the update rule for computing the target position in discrete time as in the proposed method.

\subsection{Overview of the proposed framework}
The overall framework is summarized in Algorithm~\ref{alg:cac_adaptive}.
\begin{algorithm}[tb]
    \caption{Control-Aware Command with Adaptive Execution Speed}
    \label{alg:cac_adaptive}
    \begin{algorithmic} 
        \STATE {\bfseries Input:} trained model $\pi_{\vect{\theta}}$, the horizon of the task $T_{\text{max}}$
        \STATE Set $t=0$
        \WHILE{$t < T_{\text{max}}$}
            \STATE Observe the environment
            \STATE Set $q^{\text{des}}_t$ using the actions inferred by a model $\pi_{\vect{\theta}}$
            \STATE Compute $q^{\text{targ}}_t$ with Eq.~\eqref{eq:cac}
            \STATE Evaluate the variance of the action prediction $\text{Var}(\vect{a}_t)$
            \STATE Send the command to the robot
            \IF{$\text{Var}(\vect{a}_{t}) < c_{\text{var}}$}
                \STATE $t \leftarrow t+2$
            \ELSE
                \STATE $t \leftarrow t+1$
            \ENDIF
        \ENDWHILE
    \end{algorithmic}
\end{algorithm}

The hyperparameter $c_{\text{var}}$ was fixed to $c_{\text{var}}=0.2$ throughout all experiments.

\section{Experiment Details}

\subsection{Task details in the experiment}
In the real robot experiment, the following three tasks, shown in Fig.~\ref{fig:exp_tasks}, were used to evaluate the control schemes.
\begin{enumerate}
\item \textbf{``case stacking''}  
The right arm picks up the red case and places it next to another red box. Then, the left arm picks up the black case and places it on top of the red cases.  
Episode length: 800 steps.  
Training: 50 demonstrations.  
Evaluation: 10 trials for each of three random seeds.  

\item \textbf{``box handover''}  
The right arm picks up a box and passes it to the left arm.  
Episode length: 500 steps.  
Training: 50 demonstrations.  
Evaluation: 10 trials for each of three random seeds.

\item \textbf{``pen storage''}  
The right arm opens a drawer, and the left arm grasps a pen and passes it to the right arm. Then, the right arm places the pen in the drawer and closes it.  
Episode length: 1200 steps.  
Training: 50 demonstrations.  
Evaluation: 10 trials for each of three random seeds.  
\end{enumerate}

\subsection{Other details}
The experiments with ACT was conducted based on the authors' implementation\footnote{\url{https://github.com/tonyzhaozh/act}}
The model obtained through OpenVLA-OFT was evaluated using the authors’ implementation\footnote{\url{https://github.com/moojink/openvla-oft}}, in which inference and robot control were executed sequentially, and results were compared with and without CAC. 
In addition, we conducted experiments with our alternative implementation, in which inference and robot control were executed asynchronously and in parallel through multithreading when CAC was applied.
It is worth noting that our implementation is simple and can be integrated by replacing a single file responsible for action inference and robot control in the ACT or OpenVLA-OFT repositories.
The implementation of the proposed framework is available at \url{https://github.com/TakaOsa/ACT_Async}.

In the real-robot experiments, computations were performed on a workstation equipped with an Intel Xeon w5-2455X CPU and an NVIDIA RTX A5000 GPU.



\end{document}